# Entropy, neutro-entropy and anti-entropy for neutrosophic information

**Vasile Patrascu**

Tarom Information Technology, Bucharest, Romania
e-mail: patrascu.v@gmail.com

ABSTRACT

*This article shows a deca-valued representation of neutrosophic information in which are defined the following features: truth, falsity, weak truth, weak falsity, ignorance, contradiction, saturation, neutrality, ambiguity and hesitation. Using these features, there are constructed computing formulas for entropy, neutro-entropy and anti-entropy.*

## 1. INTRODUCTION

The neutrosophic representation of information was proposed by Florentin Smarandache [11-19] and represents a generalization for the fuzzy representation proposed by Lotfi A. Zadeh [20] and in the same time, it represents an extension for intuitionistic fuzzy one proposed by Krassimir T. Atanassov [1-2]. The neutrosophic representation is defined by three parameters: degree of truth $\mu$, degree of falsity $\nu$ and degree of indeterminacy or neutrality $\omega$. In this paper, we present two deca-valued representations for neutrosophic information. There will be shown computing formulas for the following ten features of neutrosophic information: truth, falsity, weak truth, weak falsity, ignorance, contradiction, saturation, neutrality, ambiguity and hesitation. With these features we will then construct the entropy, the neutro-entropy and the anti-entropy. Further, the paper has the following structure: Section 2 presents two variants for penta-valued representation of bifuzzy information. These representations are later developed in two variants for deca-valued representation of neutrosophic information. Also, there are presented formulae for bifuzzy entropy and non-entropy. Section 3 presents two deca-valued representation of neutrosophic information and underlines three concepts for neutrosophic information: entropy, neutro-entropy and anti-entropy. If the entropy and non-entropy are already known, the neutro-entropy and anti-entropy are new concepts and were defined as sub-components of the non-entropy. For this definition there were used four components of the neutrosophic information, namely the pair weak truth with weak falsity and the pair truth with falsity. In other words, non-entropy is divided in anti-entropy and neutro-entropy. Section 4 presents some conclusions while the section 5 is the references section.

## 2. THE PENTA-VALUED REPRESENTATION OF BIFUZZY INFORMATION

The bifuzzy information is defined by the degree of truth $\mu$ and degree of falsity $\nu$. This is the primary representation. Starting from the primary representation, we can construct other derived forms [9, 10]. In the next we will present two variants.

### 2.1 Variant (I) for penta-valued representation of bifuzzy information

For the penta-valued construction, we need to define two auxiliary parameters.

The net truth:
$$\tau = \mu - \nu \qquad (1)$$

The definedness:
$$\delta = \mu + \nu - 1 \qquad (2)$$

In the next we will define the main indexes.

The bifuzzy index of ignorance:
$$\pi = max(-\delta, 0) \qquad (3)$$

The bifuzzy index of contradiction:
$$\kappa = max(\delta, 0) \qquad (4)$$

The bifuzzy index of ambiguity:
$$\alpha = 1 - |\tau| - |\delta| \qquad (5)$$

The bifuzzy index of truth:
$$\tau^+ = max(\tau, 0) \qquad (6)$$

The bifuzzy index of falsity:
$$\tau^- = max(-\tau, 0) \qquad (7)$$

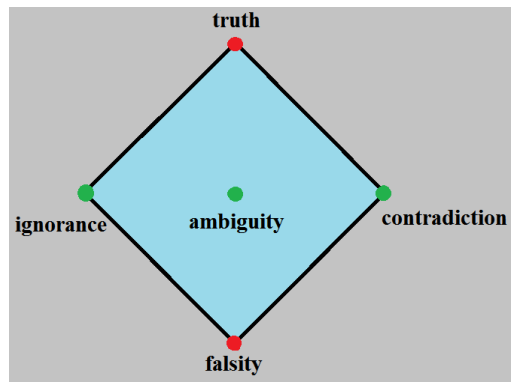

**Fig. 1. The five features of bifuzzy information**.

On this way, we obtained the first variant for penta-valued representation of bifuzzy information. There exist the following equalities:

$$\pi \cdot \kappa = 0 \qquad (8)$$
$$\tau^+ \cdot \tau^- = 0 \qquad (9)$$

The five indexes defined by formulae (3), (4), (5), (6), (7) verify the condition of partition of unity, namely:

$$\tau^+ + \tau^- + \alpha + \pi + \kappa = 1 \qquad (10)$$

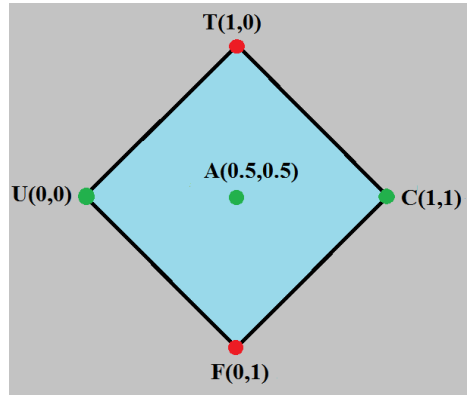

**Fig. 2. The five prototypes of bifuzzy information.**

From (10) it results the bifuzzy entropy (uncertainty) and bifuzzy non-entropy (certainty) formulae.

The bifuzzy entropy:

$$entropy = \alpha + \pi + \kappa \qquad (11)$$

The ambiguity, ignorance and contradiction are components of entropy (see figure 1).
The non-entropy is obtained by negation of entropy, namely:

$$non\_entropy = 1 - entropy \qquad (12)$$

From (10), (11) and (12) it results that the bifuzzy non-entropy is defined by:

$$non\_entropy = \tau^+ + \tau^- \qquad (13)$$

The truth and falsity are components of the non-entropy (certainty). The graphic of the constructed structure can be seen in figure 3.

The formulae (3), (4), (5), (6), (7) represent the transformation of the primary space into a penta-valued one. The next formulae defined the inverse transform from the penta-valued space to the bivalued one $(\mu, \nu)$.

$$\mu = \tau^+ + \kappa + \frac{\alpha}{2} \qquad (14)$$

$$\nu = \tau^- + \kappa + \frac{\alpha}{2} \qquad (15)$$

The two formulae (14) and (15) are equivalent with the following:

$$\begin{bmatrix}\mu\\\nu\end{bmatrix} = \tau^+ \begin{bmatrix}1\\0\end{bmatrix} + \tau^- \begin{bmatrix}0\\1\end{bmatrix} + \kappa \begin{bmatrix}1\\1\end{bmatrix} + \alpha \begin{bmatrix}0.5\\0.5\end{bmatrix} + \pi \begin{bmatrix}0\\0\end{bmatrix} \qquad (16)$$

$$\begin{bmatrix}\mu\\\nu\end{bmatrix} = \tau^+ \cdot T + \tau^- \cdot F + \kappa \cdot C + \alpha \cdot A + \pi \cdot U \qquad (17)$$

where $T, F, C, A, U$ are the prototypes shown in figure 2.

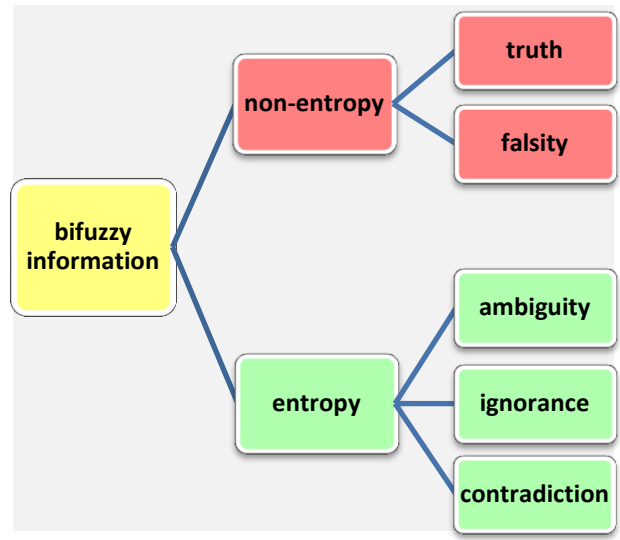

**Fig. 3. The structure of the bifuzzy information.**

## 2.2 Variant (II) for penta-valued representation of bifuzzy information

We will trace the same steps that we have used for the variant (I). We will use the same two auxiliary parameters:

The net truth:
$$\tau = \mu - \nu \tag{18}$$

The definedness:
$$\delta = \mu + \nu - 1 \tag{19}$$

In the next, we will define the main indexes.
The bifuzzy index of ignorance:
$$\pi = max(-\delta, 0)\left(1 - \frac{|\tau|}{2}\right) \tag{20}$$

The bifuzzy index of contradiction:
$$\kappa = max(\delta, 0)\left(1 - \frac{|\tau|}{2}\right) \tag{21}$$

The bifuzzy index of ambiguity:
$$\alpha = 1 - |\tau| - |\delta| + |\tau| \cdot |\delta| = (1 - |\tau|) \cdot (1 - |\delta|) \tag{22}$$

The bifuzzy index of truth:
$$\tau^+ = max(\tau, 0)\left(1 - \frac{|\delta|}{2}\right) \tag{23}$$

The bifuzzy index of falsity:
$$\tau^- = max(-\tau, 0)\left(1 - \frac{|\delta|}{2}\right) \tag{24}$$

On this way, we obtained the second variant for penta-valued representation of bifuzzy information. There exist the following equalities:

$$\pi \cdot \kappa = 0 \quad (25)$$

$$\tau^+ \cdot \tau^- = 0 \quad (26)$$

The five indexes defined by formulae (23), (24), (22), (20) and (21) verify the condition of partition of unity, namely:

$$\tau^+\left(1 - \frac{|\delta|}{2}\right) + \tau^-\left(1 - \frac{|\delta|}{2}\right) + \alpha + \pi\left(1 - \frac{|\tau|}{2}\right) + \kappa\left(1 - \frac{|\tau|}{2}\right) = 1 \quad (27)$$

From (27) it results the bifuzzy entropy (uncertainty) and bifuzzy non-entropy (certainty) formulae.

The bifuzzy entropy:

$$entropy = \alpha + \pi\left(1 - \frac{|\tau|}{2}\right) + \kappa\left(1 - \frac{|\tau|}{2}\right) \quad (28)$$

The non-entropy is obtained by negation of entropy, namely:

$$non\_entropy = 1 - entropy \quad (29)$$

From (27), (28) and (29) it results that the bifuzzy non-entropy is defined by:

$$non\_entropy = \tau^+\left(1 - \frac{|\delta|}{2}\right) + \tau^-\left(1 - \frac{|\delta|}{2}\right) \quad (30)$$

The truth and falsity are components of the non-entropy (certainty).

The formulae (20), (21), (22), (23), (24) represent the second variant for transformation of the primary space into a penta-valued one.

### 3. THE DECA-VALUED REPRESENTATION OF NEUTROSOPHIC INFORMATION

In this section we presents a deca-valued representation of neutrosophic information having as primary source the triplet $(\mu, \omega, \nu)$. This triplet defined the degree of truth, the degree of indeterminacy and the degree of falsity. We start with the penta-valued partion of the bifuzzy information and we divide each term in a sum with two other terms. For this, we will use the following formula [9]:

$$x = x \circ \omega + x \bullet \overline{\omega}$$

where " $\circ$ " and " $\bullet$ " are two conjugate Frank t_norm [3], [9]. In this paper we will use the Godel t-norm and Lukasiewicz t-norm [4], [5], [6], namely:

$$x \circ y = \min(x, y)$$

$$x \bullet y = \max(x + y - 1, 0)$$

Having two variants for penta-valued representation of bifuzzy information, we will obtain two variants for deca-valued representation for neutrosophic one.

## 3.1 Variant (I) for deca-valued representation of neutrosophic information

In this subsection, we will use the formulae (3), (4), (5), (6) and (7) that belong to the variant (I) of the bifuzzy information representation.

We decompose the bifuzzy index of truth given by (6) into the next two terms:

$$\tau^+ = \tau^+ \circ \bar{\omega} + \tau^+ \bullet \omega$$

that is equivalent with:

$$\tau^+ = \min(\tau^+, \bar{\omega}) + \tau^+ - \min(\tau^+, \bar{\omega}) \tag{31}$$

and we obtain:

- the neutrosophic index of truth

$$t = \min(\tau^+, \bar{\omega}) \tag{32}$$

- the neutrosophic index of weak truth

$$t_w = \tau^+ - \min(\tau^+, \bar{\omega}) \tag{33}$$

We decompose the bifuzzy index of falsity given by (7) into the next two terms:

$$\tau^- = \tau^- \circ \bar{\omega} + \tau^- \bullet \omega$$

that is equivalent with:

$$\tau^- = \min(\tau^-, \bar{\omega}) + \tau^- - \min(\tau^-, \bar{\omega}) \tag{34}$$

and we obtain:

- the neutrosophic index of falsity:

$$f = \min(\tau^-, \bar{\omega}) \tag{35}$$

- the neutrosophic index of weak falsity

$$f_w = \tau^- - \min(\tau^-, \bar{\omega}) \tag{36}$$

where

$$\bar{\omega} = 1 - \omega \tag{37}$$

We decompose the bifuzzy index of ignorance given by (3) into the next two terms:

$$\pi = \pi \circ \omega + \pi \bullet \bar{\omega}$$

that is equivalent with:

$$\pi = \min(\pi, \omega) + \pi - \min(\pi, \omega) \tag{38}$$

and we obtain:

- the neutrosophic index of neutrality

$$n = \min(\pi, \omega) \tag{39}$$

- the neutrosophic index of ignorance

$$u = \pi - \min(\pi, \omega) \tag{40}$$

We decompose the bifuzzy index of contradiction given by (4) into the next two terms:

$$\kappa = \kappa \circ \omega + \kappa \bullet \overline{\omega}$$

that is equivalent with:

$$\kappa = \min(\kappa, \omega) + \kappa - \min(\kappa, \omega) \tag{41}$$

and we obtain:

- the neutrosophic index of saturation

$$s = \min(\kappa, \omega) \tag{42}$$

- the neutrosophic index of contradiction

$$c = \kappa - \min(\kappa, \omega) \tag{43}$$

We decompose the bifuzy index of ambiguity into the two terms using the formula (5), namely:

$$\alpha = 2 - |\tau| - \overline{\omega} - |\delta| - \omega \tag{44}$$

It results, immediately:

$$\alpha = 2 - \min(|\tau|, \overline{\omega}) - \max(|\tau|, \overline{\omega}) - \min(|\delta|, \omega) - \max(|\delta|, \omega) \tag{45}$$

and we obtain:

- the neutrosophic index of ambiguity

$$a = 1 - \min(|\tau|, \overline{\omega}) - \max(|\delta|, \omega) \tag{46}$$

- the neutrosophic index of hesitation

$$h = 1 - \max(|\tau|, \overline{\omega}) - \min(|\delta|, \omega) \tag{47}$$

For ambiguity and hesitation we have the following equivalent formulae:

$$a = \min(\alpha + |\tau|, \overline{\omega}) - \min(|\tau|, \overline{\omega})$$

$$h = \min(\alpha + |\delta|, \omega) - \min(|\delta|, \omega)$$

Finally, we constructed the first variant for deca-valued representation for neutrosophic information. The ten parameters define a fuzzy partition of unity, namely:

$$t + t_w + f + f_w + c + u + n + s + a + h = 1 \tag{48}$$

After how we constructed the ten indexes we learn there exist the following relations:

$$t + t_w + f + f_w = |\tau| \tag{49}$$

$$c + u + n + s = |\delta| \tag{50}$$

$$a + h = \alpha \tag{51}$$

$$(t + t_w) \cdot (f + f_w) = 0 \tag{52}$$

$$(u + n) \cdot (c + s) = 0 \tag{53}$$

From the ten parameters, only four of them can be different from zero while at least six of them are zero. The formulae (32), (33), (35), (36), (39), (40) (42), (43), (46), (47) define the first variant of the transformation from the ternary space to the deca-valued one. The next formulae define the inverse transform:

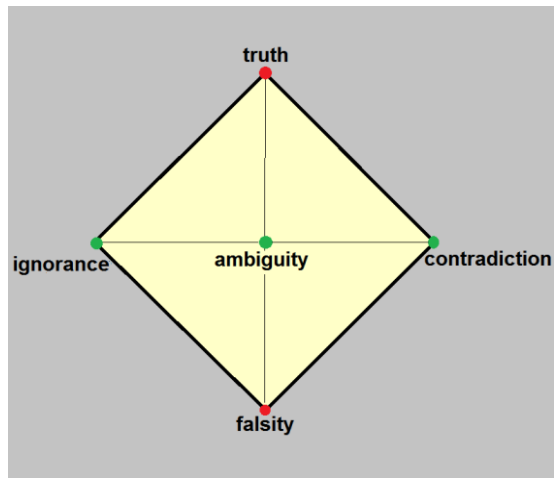

**Fig. 4. The bottom square of the neutrosophic cube.**

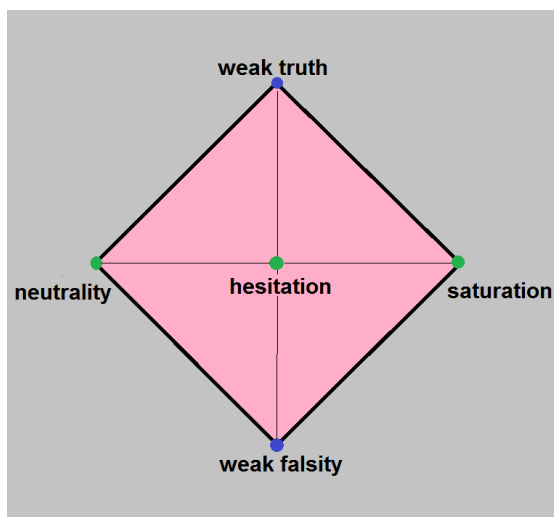

**Fig. 5. The top square of the neutrosophic cube.**

$$\mu = t + t_w + c + s + \frac{a}{2} + \frac{h}{2} \tag{54}$$

$$\omega = n + t_w + f_w + s + h \tag{55}$$

$$\nu = f + f_w + c + s + \frac{a}{2} + \frac{h}{2} \tag{56}$$

There exist the following equivalent formulae:

$$\begin{bmatrix}\mu\\\omega\\\nu\end{bmatrix} = t\begin{bmatrix}1\\0\\0\end{bmatrix} + t_w\begin{bmatrix}1\\1\\0\end{bmatrix} + f\begin{bmatrix}0\\0\\1\end{bmatrix} + f_w\begin{bmatrix}0\\1\\1\end{bmatrix} + c\begin{bmatrix}1\\0\\1\end{bmatrix} + n\begin{bmatrix}0\\1\\0\end{bmatrix} + s\begin{bmatrix}1\\1\\1\end{bmatrix} + a\begin{bmatrix}0.5\\0\\0.5\end{bmatrix} + h\begin{bmatrix}0.5\\1\\0.5\end{bmatrix}$$
$$+ u\begin{bmatrix}0\\0\\0\end{bmatrix} \tag{57}$$

$$\begin{bmatrix}\mu\\\omega\\\nu\end{bmatrix} = t \cdot \mathrm{T} + t_w \cdot \mathrm{T}_W + f \cdot \mathrm{F} + f_w \cdot \mathrm{F}_W + c \cdot \mathrm{C} + n \cdot \mathrm{N} + s \cdot \mathrm{S} + a \cdot \mathrm{A} + h \cdot \mathrm{H} +$$
$$+ u \cdot \mathrm{U} \tag{58}$$

where $T, T_W, F, F_W, C, N, S, A, H, U$ are the prototypes that can be seen in figure 7.

Forwards, as for the bifuzzy sets, firstly, we identify among the ten components, those related to uncertainty: ignorance $u$, contradiction $c$, neutrality $n$, saturation $s$, ambiguity $a$ and hesitation $h$ (see figures 4, 5, 6). These are the components of the neutrosophic uncertainty, and, in other words, the components of the neutrosophic entropy. Hence, it results the following formula for neutrosophic entropy calculation:

$$entropy = u + c + n + s + a + h \qquad (59)$$

Further, as for the bifuzzy sets, the negation of entropy leads to the neutrosophic non-entropy, namely:

$$non\_entropy = 1 - entropy \qquad (60)$$

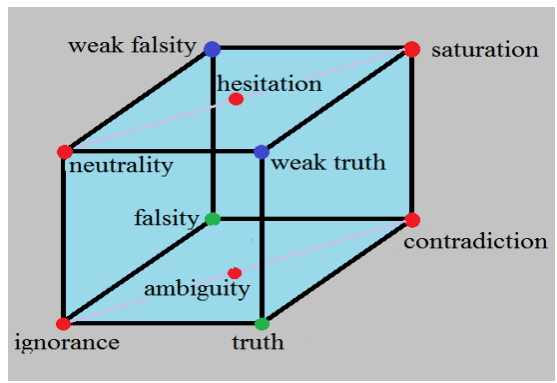

Fig. 6. The neutrosophic cube and the ten features distribution.

From (48), (59) and (60) it results the following formula for non-entropy calculation:

$$non\_entropy = t + f + t_w + f_w \qquad (61)$$

Now, analyzing formula (61), it is seen that non-entropy is not identified with the neutrosophic certainty because weak truth and weak falsity cannot be components of certainty. These two components, $t_w$ and $f_w$ not belong to certainty and in the same time not belong to uncertainty. These components are found somewhere between certainty and uncertainty, namely in a middle zone. In fact, the two components define a new entity, the neutro-entropy or simply neutropy:

$$neutro\_entropy = t_w + f_w \qquad (62)$$

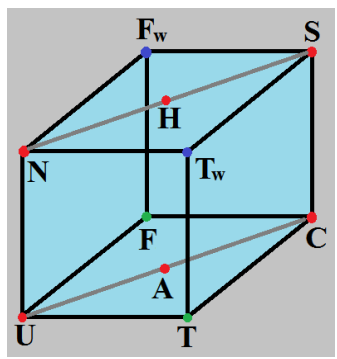

Fig. 7. The neutrosophic cube and the ten prototypes distribution.

Consequently, the only certainty components are $t$ and $f$ and in other words, these will be components of the neutrosophic anti-entropy.

Vasile Patrascu, Technical Report, TI.1.6.2017.

$$anti\_entropy = t + f \qquad (63)$$

Finally, we come to decompose the non-entropy in two parts, non-entropy and anti-entropy, existing the next formula:

$$non\_entropy = anti\_entropy + neutro\_entropy \qquad (64)$$

In figure 8, we can see the detailed structure of the neutrosophic information.

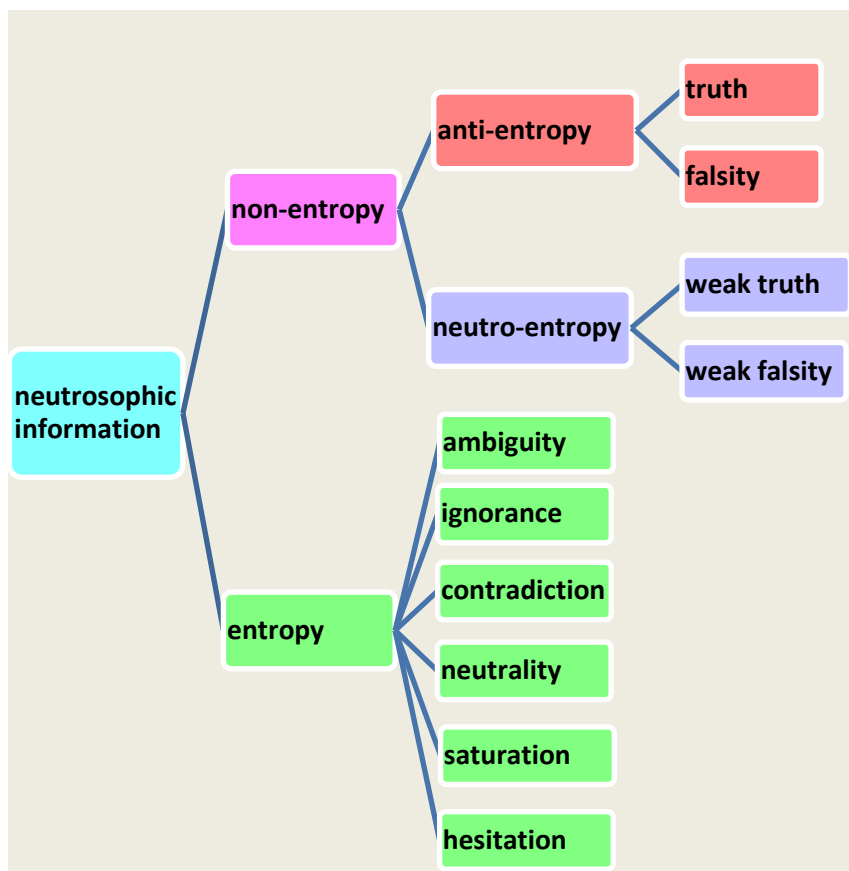

**Fig. 8. The structure of the neutrosophic information**

Once again, it highlights the principle laid down by Smarandache, principle that led to the development theory of neutrosophy, namely, for each entity $A$ there is an opposite entity $anti\_A$ and between these two, there exists the third entity, more precisely, one situated in the middle, $neut\_A$. In addition, $anti\_A$ and $neut\_A$ together form $non\_A$. In the particular case of the neutrosophic information associated with the decomposition given by formula (48), among these three components exists the partition of unity, namely:

$$entropy + neutro\_entropy + anti\_entropy = 1 \qquad (65)$$

### 3.3 Variant (II) for deca-valued representation of neutrosophic information

In this subsection we will use the formulae (20), (21), (22), (23) and (24) that belong to the variant (II) of bifuzzy information representation.
We decompose the bifuzzy index of truth, given by (23) into the next two terms:

$$\tau^+ = \tau^+ \circ \overline{\omega} + \tau^+ \bullet \omega$$

that is equivalent with:

$$\tau^+ = \min(\tau^+, \overline{\omega}) + \tau^+ - \min(\tau^+, \overline{\omega}) \tag{66}$$

and we obtain:

- the neutrosophic index of truth

$$t = \min(\tau^+, \overline{\omega})\left(1 - \frac{|\delta|}{2}\right) \tag{67}$$

- the neutrosophic index of weak truth

$$t_w = \left(\tau^+ - \min(\tau^+, \overline{\omega})\right)\left(1 - \frac{|\delta|}{2}\right) \tag{68}$$

We decompose the bifuzzy index of falsity given by (24) into the next two terms:

$$\tau^- = \tau^- \circ \overline{\omega} + \tau^- \bullet \omega$$

that is equivalent with:

$$\tau^- = \min(\tau^-, \overline{\omega}) + \tau^- - \min(\tau^-, \overline{\omega}) \tag{69}$$

and we obtain:

- the neutrosophic index of falsity:

$$f = \min(\tau^-, \overline{\omega})\left(1 - \frac{|\delta|}{2}\right) \tag{70}$$

- the neutrosophic index of weak falsity

$$f_w = \left(\tau^- - \min(\tau^-, \overline{\omega})\right)\left(1 - \frac{|\delta|}{2}\right) \tag{71}$$

where

$$\overline{\omega} = 1 - \omega \tag{72}$$

We decompose the bifuzzy index of ignorance given by (20) into the next two terms:

$$\pi = \pi \circ \omega + \pi \bullet \overline{\omega}$$

that is equivalent with:

$$\pi = \min(\pi, \omega) + \pi - \min(\pi, \omega) \tag{73}$$

and we obtain:

- the neutrosophic index of neutrality

$$n = \min(\pi, \omega)\left(1 - \frac{|\tau|}{2}\right) \tag{74}$$

- the neutrosophic index of ignorance

$$u = \bigl(\pi - \min(\pi, \omega)\bigr)\left(1 - \frac{|\tau|}{2}\right) \tag{75}$$

We decompose the bifuzzy index of contradiction given by (21) into the next two terms:

$$\kappa = \kappa \circ \omega + \kappa \bullet \overline{\omega}$$

that is equivalent with:

$$\kappa = \min(\kappa, \omega) + \kappa - \min(\kappa, \omega) \tag{76}$$

and we obtain:

- the neutrosophic index of saturation

$$s = \min(\kappa, \omega)\left(1 - \frac{|\tau|}{2}\right) \tag{77}$$

- the neutrosophic index of contradiction

$$c = \bigl(\kappa - \min(\kappa, \omega)\bigr)\left(1 - \frac{|\tau|}{2}\right) \tag{78}$$

We decompose the bifuzy index of ambiguity given by (22) into two terms. Firstly, using formula (22) we obtain the following equivalent form:

$$\alpha = 2 - |\tau|\left(1 - \frac{|\delta|}{2}\right) - \overline{\omega} - |\delta|\left(1 - \frac{|\tau|}{2}\right) - \omega \tag{79}$$

Then, it results, immediately:

$$\alpha = 2 - \min\!\left(|\tau|\left(1 - \frac{|\delta|}{2}\right), \overline{\omega}\right) - \max\!\left(|\tau|\left(1 - \frac{|\delta|}{2}\right), \overline{\omega}\right) - \min\!\left(|\delta|\left(1 - \frac{|\tau|}{2}\right), \omega\right)$$
$$- \max\!\left(|\delta|\left(1 - \frac{|\tau|}{2}\right), \omega\right) \tag{80}$$

and we obtain:

- the neutrosophic index of ambiguity

$$a = 1 - \min\!\left(|\tau|\left(1 - \frac{|\delta|}{2}\right), \overline{\omega}\right) - \max\!\left(|\delta|\left(1 - \frac{|\tau|}{2}\right), \omega\right) \tag{81}$$

- the neutrosophic index of hesitation

$$h = 1 - \max\!\left(|\tau|\left(1 - \frac{|\delta|}{2}\right), \overline{\omega}\right) - \min\!\left(|\delta|\left(1 - \frac{|\tau|}{2}\right), \omega\right) \tag{82}$$

For ambiguity and hesitation we have the following equivalent formulae:

$$a = \min\!\left(\alpha + |\tau|\left(1 - \frac{|\delta|}{2}\right), \overline{\omega}\right) - \min\!\left(|\tau|\left(1 - \frac{|\delta|}{2}\right), \overline{\omega}\right)$$

$$h = \min\!\left(\alpha + |\delta|\left(1 - \frac{|\tau|}{2}\right), \omega\right) - \min\!\left(|\delta|\left(1 - \frac{|\tau|}{2}\right), \omega\right)$$

Finally, we constructed the second variant for deca-valued representation of neutrosophic information. The ten parameters define a fuzzy partition of unity.

$$t + t_w + f + f_w + c + u + n + s + a + h = 1 \tag{83}$$

After how we constructed the ten indexes we learn there exist the following relations:

$$t + t_w + f + f_w = |\tau|\left(1 - \frac{|\delta|}{2}\right) \tag{84}$$

$$c + u + n + s = |\delta|\left(1 - \frac{|\tau|}{2}\right) \tag{85}$$

$$a + h = \alpha \tag{86}$$

$$(t + t_w) \cdot (f + f_w) = 0 \tag{87}$$

$$(u + n) \cdot (c + s) = 0 \tag{88}$$

From the ten parameters only four of them can be different from zero while at least six of them are zero.

The formulae (67), (68), (70), (71), (74), (75) (77), (78), (81), (82) define another transformation from the ternary primary space to the deca-valued one. Hence, it results the following formulae for entropy, non-entropy, neutro-entropy and anti-entropy:

$$entropy = 1 - |\tau|\left(1 - \frac{|\delta|}{2}\right) \tag{89}$$

$$non\_entropy = |\tau|\left(1 - \frac{|\delta|}{2}\right) \tag{90}$$

$$neutro\_entropy = (|\tau| - \min(|\tau|, \bar{\omega}))\left(1 - \frac{|\delta|}{2}\right) \tag{91}$$

$$anti\_entropy = \min(|\tau|, \bar{\omega})\left(1 - \frac{|\delta|}{2}\right) \tag{92}$$

And again there exist the following relation:

$$non\_entropy = anti\_entropy + neutro\_entropy \tag{93}$$

$$entropy + neutro\_entropy + anti\_entropy = 1 \tag{94}$$

## 4. CONCLUSION

This approach presents a multi-valued representation of the neutrosophic information. It highlights the link between the bifuzzy information and neutrosophic one. The constructed deca-valued structures show the neutrosophic information complexity. These deca-valued structures led to construction of two new concepts for the neutrosophic information: neutro-entropy and anti-entropy. These two concepts are added to the two existing: entropy and non-entropy. Thus, we obtained the following triad: entropy, neutro-entropy and anti-entropy. For the moment, neutro-entropy was defined for neutrosophic information but it is possible that in the future this concept will be defined for other research fields such as biology. For now, in biology, it was defined anti-entropy and from the start was stated that this is different from non-entropy [21], [22]. In the future, we will see if perhaps what differentiates them is just neutro-entropy.